\documentclass[11pt,a4paper]{article}
\usepackage[hyperref]{emnlp-ijcnlp-2019}
\usepackage{times}
\usepackage{latexsym}
\usepackage{url}
\usepackage{lipsum}
\usepackage{amsmath}
\usepackage{amssymb}
\usepackage{multirow}
\usepackage{bbm}
\usepackage{titlecaps}
\usepackage{booktabs}
\usepackage{graphicx}
\usepackage{xcolor,colortbl}
\usepackage{tikz}
\usepackage{soul}

\newcommand{\rom}[1]{\uppercase\expandafter{\romannumeral #1\relax}}

\definecolor{Gray}{gray}{0.85}
\definecolor{LightCyan}{rgb}{0.88,1,1}
\newcolumntype{a}{>{\columncolor{Gray}}c}
\newcolumntype{b}{>{\columncolor{white}}c}

\graphicspath{ { ./figures/} }

\newcommand{\Ev}{\mathbf{E}}

\newcommand{\Hv}{\mathbf{H}}

\newcommand{\Mv}{\mathbf{M}}

\newcommand{\Vv}{\mathbf{V}}
\newcommand{\Wv}{\mathbf{W}}

\aclfinalcopy 

\title{Towards Zero-resource Cross-lingual Entity Linking}

\author{Shuyan Zhou, Shruti Rijhwani, Graham Neubig\\\\
  Language Technologies Institute \\
  Carnegie Mellon University \\
  \texttt{\{shuyanzh,srijhwan,gneubig\}@cs.cmu.edu} \\
}

\date{}

\begin{document}
\maketitle
\begin{abstract}

Cross-lingual entity linking (XEL) grounds named entities in a source language to an English Knowledge Base (KB), such as Wikipedia. XEL is challenging for most languages because of limited availability of requisite resources. However, much previous work on XEL has been on simulated settings that actually use significant resources (e.g. source language Wikipedia, bilingual entity maps, multilingual embeddings) that are unavailable in truly low-resource languages. In this work, we first examine the effect of these resource assumptions and quantify how much the availability of these resource affects overall quality of existing XEL systems. Next, we propose three improvements to both entity candidate generation and disambiguation that make better use of the limited data we do have in resource-scarce scenarios. With experiments on four extremely low-resource languages, we show that our model results in gains of 6-23\% in end-to-end linking accuracy.%
\footnote{Code is available at \url{https://github.com/shuyanzhou/burn_xel}}
\end{abstract}

\section{Introduction}\label{introduction}


Entity linking (EL;  \citet{bunescu2006using,cucerzan2007large,dredze2010entity,hoffart2011robust}) identifies entity mentions in a document and associates them with their corresponding entries in a structured Knowledge Base (KB)~\cite{shen2015entity}, such as Wikipedia or Freebase~\cite{bollacker2008freebase}.
EL involves two main steps: (1) \emph{candidate generation}, retrieving a list of candidate KB entries for each entity mention, and (2) \emph{disambiguation}, selecting the most likely entry from the candidate list.

\begin{figure}
    \centering
    \includegraphics[width=\columnwidth]{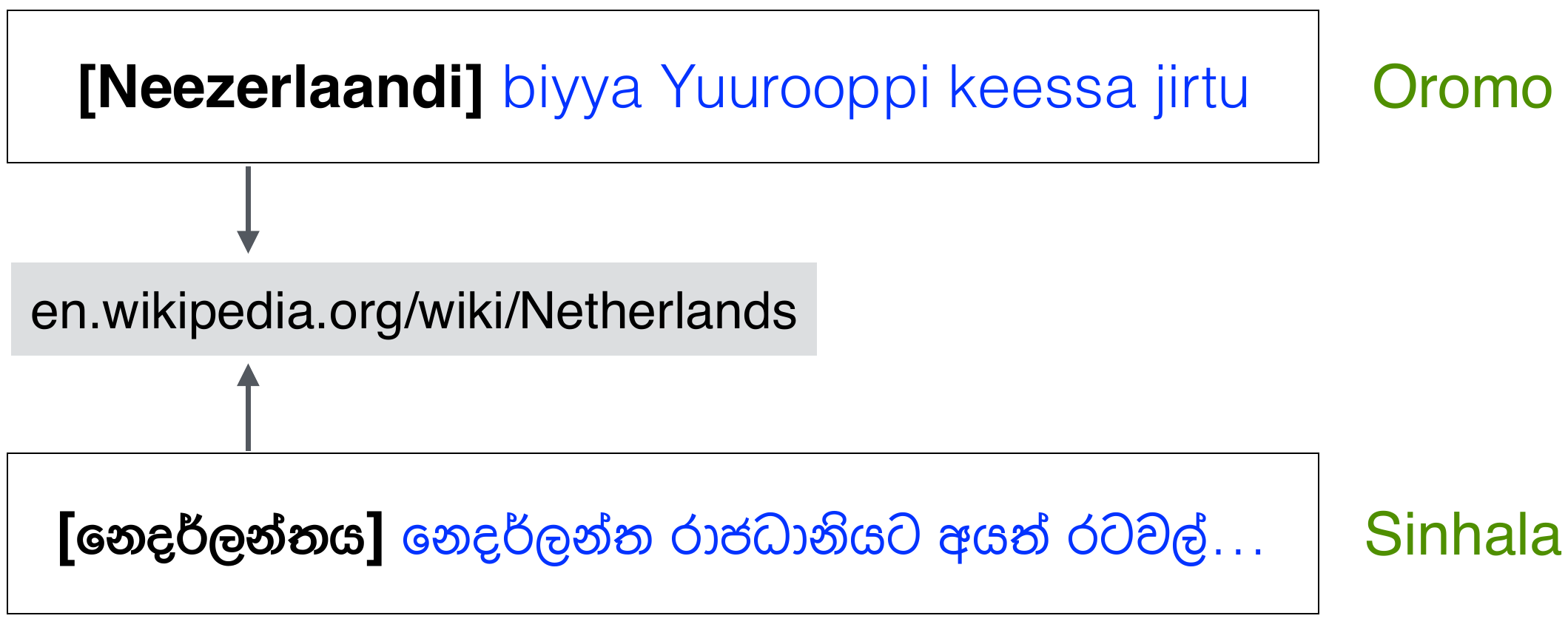}
    \vspace{-7mm}
    \caption{XEL for two low-resource languages -- Oromo and Sinhala, linking source mentions to entity ``Netherlands'' in English Wikipedia.
    }
    \label{fig:example}
    \vspace{-5mm}
\end{figure}

In this work, we focus on cross-lingual entity linking (XEL; \citet{mcnamee2011cross}, \citet{ji2015overview}), where the document is in a (source) language that is different from the (target) language of the KB. Following recent work~\cite{sil2018neural, upadhyay2018joint}, we use English Wikipedia as this KB. Figure \ref{fig:example} shows an example.

XEL to English from major languages such Spanish and Chinese has been carefully studied, and significant progress has been made.
Success in these languages can be largely attributed to the availability of rich resources.
Specifically, the following is a list of resources required by recent works~\cite{tsai-roth:2016:N16-1,pan2017cross,sil2018neural,upadhyay2018joint}:

\noindent
\textbf{English Wikipedia} ($\mathbbm{W}_{\textrm{eng}}$): The target KB and a large corpus of text.
Importantly, the text is annotated with anchor text linking between entity mentions (e.g. ``Holland'' in the body text of an article) and the page for the entity (e.g. ``Netherlands'').
These annotations can be used to extract mention-entity maps for entity candidate generation, and to directly train entity disambiguation systems.

\noindent
\textbf{Source Language Wikipedia} ($\mathbbm{W}_{\textrm{src}}$):
KB and corresponding text in the source language.
Similarly to English Wikipedia, this can be used to obtain mention-entity maps or train disambiguation systems, but the size of Wikipedia is relatively small for most low-resource languages.

\noindent
\textbf{Bilingual Entity Maps} ($\mathbbm{M}$): A map between source language entities and English entities.
One common source of this map is Wikipedia inter-language links between the source language and English. These inter-language links can directly and unambiguously link entities in the source language KB to the English KB.

\noindent
\textbf{Multilingual Embeddings} ($\mathbbm{E}$):
These embeddings map words in different languages to the same vector space.

The availability of these resources varies widely among languages.
They are available for high-resource languages such as Spanish and Chinese, which have been widely used as test-beds for XEL.
For example, there are over 1.5 million articles in Spanish Wikipedia, which provide an abundance of annotations.
However, the situation is not as favorable for most other languages: while $\mathbbm{W}_{\textrm{eng}}$ is invariant of the source language to link from, many of the other resources are small or non-existent.
In fact, only 300 languages (from $\approx$7000 living languages in the world) have Wikipedia $\mathbbm{W}_{\textrm{src}}$, and among these many have a limited number of pages. For example, Oromo, a Cushitic language with 30 million speakers, has only 776 Wikipedia pages.
It is similarly difficult to obtain exhaustive bilingual entity maps, and for many languages even the monolingual/parallel text necessary to train multilingual embeddings is scarce.


This work makes two major contributions regarding XEL for low-resource languages.

The first major contribution is empirical.
We extensively evaluate the effect of resource restrictions on existing XEL methods in true low-resource settings instead of simulated ones (Section \ref{base_experiment}). We compare the performance of both the candidate generation model and the disambiguation model of our baseline XEL system between two high-resource languages and four low-resource languages. We quantify how much the availability of the aforementioned resources affect the overall quality of the existing methods, and find that \emph{with scarce access to these resources, the performance of existing methods drops significantly}. This highlights the effect of resource constraints in realistic settings, and indicates that these constraints should be considered more carefully in future system design.

Our second major contribution is methodological.
We propose three methods as first steps towards ameliorating the large degradation in performance we see in low-resource settings.
(1) We investigate a \emph{hybrid candidate generation method}, combining existing lookup-based and neural candidate generation methods to improve candidate list recall by 9-24\%.
(2) We propose a set of \emph{entity disambiguation features that are entirely language-agnostic}, allowing us to train a disambiguation system on English and transfer it directly to low-resource languages.
(3) We design a \emph{non-linear feature combination} method, which makes it possible to combine features in a more flexible way.
We test these three methodological improvements on four extremely low-resource languages (Oromo, Tigrinya, Kinyarwanda, and Sinhala), and find that the combination of these three techniques leads to consistent performance gains in all four languages, amounting to 6-23\% improvement in end-to-end XEL accuracy.

\section{Problem Formulation}\label{formulation}
Given a set of documents $\mathcal{D}=\{D_1, D_2, ..., D_l\}$ in any source language $L_s$, a set of detected mentions $\Mv_D=\{m_1, m_2, ..., m_n\}$ for each document $D$, and the English Wikipedia $\Ev_{\textrm{KB}}$, the goal of XEL is to associate each mention with its corresponding entity in the English Wikipedia. We denote an entity in English Wikipedia as $e$ and its parallel entity in the source language Wikipedia as $e^\textrm{src}$.

For each $m_i \in \Mv_{D}$, candidate generation first retrieves a list of candidate entities $\mathbf{e}_{i}=\{e_{i,1}, e_{i, 2}, ..., e_{i, n}\}$ from $\Ev_{\textrm{KB}}$ based on probabilities $\mathbf{p}_{i}=\{p_{i,1}, p_{i, 2}, ..., p_{i, n}\}$ where $p_{i, j}$ denotes $p(e_{i,j}|m_i)$. Then, the disambiguation model assigns a score $s(e_{i,j}|D)$ to each $e_{i,j}$. These scores are normalized among $\mathbf{e}_i$ and result in the probability $p(e_{i,j}|D)$. The entity with highest score is selected as the prediction. We denote the gold entity as $e^{*}$.

Performance of candidate generation is measured by \emph{gold candidate recall}: the proportion of mentions whose top-$n$ candidate list contains the gold entity over all test mentions. This recall upper-bounds performance of an entity disambiguation system. In the consideration of the computational cost of the more complicated downstream disambiguation model, this $n$ is often 30 or smaller~\cite{sil2018neural, upadhyay2018joint}. The performance of an end-to-end XEL system is measured by \emph{accuracy}: the proportion of mentions whose predictions are correct. We follow \citet{yamada2017learning,ganea2017deep} and focus on \emph{in-KB} accuracy; we ignore mentions whose linked entity does not exist in the KB in this work.

\section{Baseline Model}\label{baseline}
This section describes existing methods for candidate generation and disambiguation, and our baseline XEL system, which is heavily inspired by existing works \cite{ling2015design, globerson2016collective, pan2017cross}. We investigate the effect of resource constraints on this system in Section \ref{base_experiment}. Based on empirical observations, we propose our improved XEL system in Section \ref{model} and present its results in Section \ref{experiment}. 

\subsection{Candidate Generation}\label{sec:candidate_generation}
\textsc{WikiMention}: With access to all the resources we list above, there is a straightforward approach to candidate generation used by most state-of-the-art work in XEL~\cite{sil2018neural,upadhyay2018joint}.
Specifically, a monolingual mention-entity map can be extracted from $\mathbbm{W}_{\textrm{src}}$ by finding all cross-article links in $\mathbbm{W}_{\textrm{src}}$, and using the anchor text as mention $m$ and the linked entity as $e^\textrm{src}$. These entities are then redirected to English Wikipedia with $\mathbbm{M}$ to obtain $e$. 
For instance, if Oromo mention ``Itoophiyaatti'' is linked to entity ``Itoophiyaa'' 
in some Oromo Wikipedia pages, the corresponding English Wikipedia entity ``Ethiopia'' will be acquired through $\mathbbm{M}$ and used as a candidate entity for the mention. The score $p(e_{i,j}|m_i)$ provided by this model shows the \emph{probability} of linking to $e_{i,j}$ when mentioning $m_{i}$. 
%
Because of its heavy reliance on  $\mathbbm{W}_{\textrm{src}}$ and $\mathbbm{M}$, \textsc{WikiMention} does not generalize well to real low-resource settings. We discuss this in Section~\ref{comparision}.


\textsc{Pivoting}: Recently,~\citet{rijhwani19aaai} propose a zero-shot transfer learning method for XEL candidate generation, which uses no resources in the source language. A character-level LSTM is trained to encode entities using a bilingual entity map between some high-resource language and English. If the chosen high-resource language is closely related to the low-resource language (same language family, shared orthography etc.), zero-shot transfer will often be successful in generating candidates for the low-resource language. In this case, the model generated score $s(e_{i,j}|m_i)$ indicates the \emph{similarity} which should be further normalized into a \emph{probability} $p(e_{i,j}|m_i)$ (Section \ref{sec:calibration}).

Notably, both methods have advantages and disadvantages, with \textsc{Pivoting} generally being more robust, and \textsc{WikiMention} being more accurate when resources are available. To take advantage of this, we propose a method for calibrated combination of these two methods in Section \ref{sec:calibration}.



\subsection{Featurization and Linear Scoring}\label{sec:base_inference}

Next, we move to the entity disambiguation step, which we further decompose into (1) the design of features and (2) the choice of inference model that combines these features together.

\subsubsection{Featurization}
Unfortunately for low-resource settings, many XEL disambiguation models rely on extensive resources such as $\mathbbm{E}$ and $\mathbbm{W}_{\textrm{src}}$ \cite{sil2018neural, upadhyay2018joint} to obtain features.
However, some previous work on XEL does limit its resource usage to $\mathbb{W}_{\text{eng}}$, which is available regardless of the source language. Our baseline follows one such method by \citet{pan2017cross}.

We use two varieties of features: \emph{unary} features that reflect properties of a single entity and \emph{binary} features that quantify coherence between pairs of entities. 
The top half of Table~\ref{tab:1} shows unary feature functions, which take one argument $e_{i,j}$ and return a value that represents some property of this entity. The grayed mention-entity prior $f_l^1(e_{i,j})$ is the main unary feature used by \citet{pan2017cross}, and we use this in our baseline. Binary features are in the bottom half of Table~\ref{tab:1}. Each binary feature function $f_g^i(e_{i,j}, e_{k, w})$ takes two entities as arguments, and returns a value that indicates the relatedness between the entities. Similarly, the grayed co-occurrence feature $f_g^1(e_{i, j}, e_{k, w})$ is used in the baseline. We refer to these two features as \textsc{Base}.

While these features have proven useful in higher-resource XEL, in lower-resource scenarios, we hypothesize that it is more important to design features that make the most use of the language-invariant resource $\mathbbm{W}_{\textrm{eng}}$ to make up for the relative lack of other resources in the source language. We discuss more intelligent features in Section~\ref{feature}. 

\subsubsection{Non-iterative Linear Inference Model}
While the design of features is resource-sensitive, the choice of an inference model is fortunately resource-agnostic as it only relies on the existence of features.
Our baseline follows existing (X)EL works~\cite{ling2015design, globerson2016collective, pan2017cross} to \emph{linearly} aggregate unary features to a \emph{local} score $s_l(e|D)$ and binary features to a \emph{global} score $s_g(e|D)$.
The local score reflects the properties of an independent entity, and the global score quantifies the coherence between an entity and other linked entities in the document.
The score of each entity is defined as:
\begin{equation}\label{burn:final}
    s(e_{i, j}|D)=s_g(e_{i, j}|D) + s_l(e_{i, j}|D)
\nonumber
\end{equation}

The local score is the linear combination of unary features $f_l^i(e_{i,j}) \in \mathbf{\Phi}(e_{i,j})$:
\begin{equation}\label{eq:base_l}
    s_l(e_{i,j}|D)={\Wv_l^T}\mathbf{\Phi}(e_{i,j})
\end{equation}
where $\textbf{W}_l \in \mathbbm{R}^{d_l \times 1}$ and $d_l$ is the number of unary features in the vector.

On the other hand, the global score $s_g$ is an average aggregation of mention evidence $s_m$ across the document. Each $s_m(m_k, e_{i,j})$ indicates how strongly a context mention $m_k$ supports the $j$-th candidate entity of mention $m_i$:
\begin{equation}\label{eq:base_g}
    s_g(e_{i,j}|D)=\frac{1}{|\Mv_{D}|}\sum_{k \neq i}s_m(m_k, e_{i,j})
\end{equation}

As a mention is in fact the surface form of other candidate entities, $s_m(m_k, e_{i,j})$ can be measured by the relatedness between the candidate entities $\mathbf{e}_k$ of $m_k$ and $e_{i,j}$.
Our baseline inference model follows~\citet{ling2015design,globerson2016collective} to process this evidence in a \textsc{Greedy} manner:
\begin{equation}\label{eq:base_m}
    s_m(m_k, e_{i,j}) = \max_{e_{k,w}\in \Ev_{k}}(s_e(e_{i, j}, e_{k, w}))
\end{equation}
Similarly to $s_l$, $s_e(e_{i, j}, e_{k, w})$ is the linear combination of binary features $f_g^i(e_{i,j}, e_{k, w}) \in \mathbf{\Psi}(e_{i, j}, e_{k, w})$:
\begin{equation}\label{eq:base_e}
    s_e(e_{i,j}, e_{k, w})={\Wv_g^T}\mathbf{\Psi}(e_{i,j}, e_{k, w})
\end{equation}

The greedy strategy often results in a sub-optimal assignment, as the confidence of each candidate entity is not taken into consideration. To solve this problem, we propose iteratively updating belief of each candidate entity in Section \ref{sec:burn}.

Following \citet{upadhyay2018joint, sil2018neural}, we consider \textsc{WikiMention} as the baseline candidate generation model and \textsc{Base}+\textsc{Greedy} as the baseline disambiguator. We denote \textsc{WikiMention}+\textsc{Base}+\textsc{Greedy} as the end-to-end baseline system.

\begin{table*}[t]
\small
  \centering
  \resizebox{\textwidth}{!}{%
  \begin{tabular}{cccc}
  \toprule
  \textbf{Symbol} & \textbf{Feature Name} & \textbf{Equation} & \textbf{Resource} \\
  \midrule
  
  \rowcolor{Gray} $f_l^1(e_{i,j})$ & Mention-entity prior score & 
  $\log(\max(p(e_{i,j}|m_i), \epsilon))$ & Variable \\
  
  $f_l^2(e_{i,j})$ & Entity prior & $\log(\max(\frac{c(e_{i,j})}{\sum_{e\in \Ev_{\textrm{KB}}} c(e)}, \epsilon))$ & $\mathbbm{W}_{\textrm{eng}}$ \\
  
  $f_l^3(e_{i,j})$ & Related mention number & $\sum_{m_k \in \Mv_{D}\backslash m_i}\mathbbm{1}(\text{any}_{e_{k,m} \in \Ev_{k}}f_g^1(e_{i,j}, e_{k,m})>0)$ & -  \\
  
  $f_l^4(e_{i,j})$ & Exact match number & $\sum_{m_{k} \in \Mv_{D}\backslash m_i}\mathbbm{1}(e\in \Ev_{k})$ & - \\
  \midrule
  \rowcolor{Gray}
  $f_g^1(e_{i,j}, e_{k,w})$ & Co-occurrence probability & $\log(\max(\frac{c(e_{i,j}, e_{k,w})}{c(e_{i,j})}), \epsilon)$ & $\mathbbm{W}_{\textrm{eng}}$ \\
  
  $f_g^2e_{i,j}, e_{k,w})$ & Positive Pointwise Mutual Information (PPMI) & $\max (\log_2 (\frac{p(e_{i,j}, e_{k,w})}{p'(e_{i,j})p'(e_{k,w})}), 0)$ & $\mathbbm{W}_{\textrm{eng}}$  \\
  
  $f_g^3(e_{i,j}, e_{k,w})$ & Entity embedding similarity & $\text{cosine}(\Vv_{e_{i,j}}, \Vv_{e_{k,w}})$ & $\mathbbm{W}_{\textrm{eng}}$ \\
  
  $f_g^4(e_{i,j}, e_{k,w})$ & Hyperlink count & $\log(\max(\frac{\sum_{e_k \in \Hv_{e_{i,j}}}\mathbbm{1}(e_{i,j}=e_{k,w})}{|\Hv_{e_{i,j}}|}, \epsilon)$ & $\mathbbm{W}_{\textrm{eng}}$ \\
  \bottomrule
  
  \end{tabular}%
  }
  \caption{Unary features (top half) and binary features (bottom half). Gray indicates \textsc{Base} features. ``Variable'' means this feature comes from the candidate generation model and thus its resource dependency will be decided by that model; $\epsilon$ is set to \texttt{1e-7}; $c(e)$ is the frequency of an entity among all anchor links in $\mathbbm{W}_{\textrm{eng}}$; $c(e_{i}, e_{j})$ is the co-occurrence count of two entities in $\mathbbm{W}_{\textrm{eng}}$; $p(e_{i}, e_{j})$ is normalized over all entity pairs and $p'(e_{i})$ is normalized over all entities with smoothing parameter $\gamma=0.75$; $\Vv_{e}$ represents the entity embedding of $e_{i}$; $\Hv_{e_{i}}$ represents a set of entities in $e_{i}$'s English Wikipedia page.}
  \label{tab:1}
\end{table*}
\section{Experiment \rom{1}: Real Low-resource Constraints in XEL}
\label{base_experiment}
In this section, we study the effects of resource constraints in truly low-resource settings; we then evaluate how this changes the conclusions we may draw about the efficacy of existing XEL models. We attempt to answer the following research questions: (1) how the does the availability of resources influence the performance of XEL systems, and (2) how do truly low-resource settings diverge from XEL with more resources?

We perform this study within the context of our \textsc{WikiMention+Base+Greedy} baseline (which is conceptually similar to previous work). We carry out the study on several languages and datasets:
\smallskip

\noindent
\textbf{TAC-KBP}: TAC-KBP 2011 for English (\textit{en}) \cite{tac2011overview}, TAC-KBP 2015 for Spanish (\textit{es}) and Chinese (\textit{zh})~\cite{ji2015overview}. All contain documents from forums and news. 
\smallskip

\noindent
\textbf{DARPA-LRL}: The DARPA LORELEI annotated documents\footnote{https://www.darpa.mil/program/low-resource-languages-for-emergent-incidents} in 4 low-resource languages: Tigrinya (\textit{ti}), Oromo (\textit{om}), Kinyarwanda (\textit{rw}) and Sinhala (\textit{si}). These are news articles, blogs and social media posts about disasters and humanitarian crises.

Detailed experimental settings are in Section \ref{train_detail}. 
It is notable that a large number of previous works examine XEL on simulated low-resource settings such as the TAC-KBP datasets for large languages such as Chinese and English \cite{sil2018neural, upadhyay2018joint}, while the DARPA-LRL datasets are more reflective of true constraints in low-resource scenarios.

\subsection{Results}\label{comparision}
\begin{table*}[h]
\small
\label{table:dataset}
\begin{center}

\begin{tabular}{cccccccc}

\toprule
& & \multicolumn{2}{c}{\emph{high-resource}} & \multicolumn{4}{c}{\emph{low-resource}} \\\cmidrule(lr){3-4} \cmidrule(lr){5-8}
Model & en & zh & es & ti & om & rw & si \\
\midrule
Gold Candidate Recall & 92.4 & 89.2 & 89.0 &21.9&  45.3&  45.6&  66.6 \\

$p(e|m)$ & 70.1 &83.1& 78.2&21.5&41.0& 45.1& 63.1\\

\textsc{Base+Greedy} & 77.5&85.5& 82.9& 21.8&  38.4&  44.9&  64.4\\



\midrule
Wikipedia Size & 5.0M & 1.0M & 1.5M & 168 & 775 & 1.8K & 15.1K \\
\bottomrule
\end{tabular}
\caption{Gold candidate recall of \textsc{WikiMention} over seven languages, accuracy (\%) of selecting the highest score entity, and accuracy after end-to-end EL using the \textsc{Base+Greedy} method.}
\label{info}
\vspace{-5mm}
\end{center}
\end{table*}


Table \ref{info} shows various statistics for the baseline system on English, two high-resource, and four low-resource XEL languages.
The first row of Table \ref{info} shows the gold candidate recall of \textsc{WikiMention} on 7 languages.
The Wikipedia sizes of each language are shown in the last row of the table for reference.
In general, the gold candidate recall of \textsc{WikiMention} is positively correlated with the size of available Wikipedia resoruces. We can note that compared to the four low-resource languages, the statistics of the two high-resource languages are closer to those of English.

End-to-end performance of a system that selects the entity with the highest score according to \textsc{WikiMention} is listed in the second row of the table. This trivial context-insensitive disambiguation method results in performance not far from the upper bound in six XEL languages. However, the size of the gap between this method and the upper bound is largely different between high- and low-resource settings -- this gap is significant for high-resource languages, but quite small for the four low-resource languages.
Accordingly, in third row where we apply the disambiguation method \textsc{Base+Greedy}, we find gains of 2-7\% on the high-resource languages, but little to no gain on the low-resource languages.
This shows that when using a standard candidate generation method such as \textsc{WikiMention}, there is \emph{little room for more sophisticated disambiguation models to improve performance}, despite the fact that development of disambiguation methods (rather than candidate generation) has been the focus of much prior work.

\section{Proposed Model Improvements} \label{model}
Next, we introduce our proposed methods: (1) calibrated combination of two existing candidate generation models, (2) an XEL disambiguation model that makes best use of resources that \emph{will} be available in extremely low-resource settings. 

\subsection{Calibrated Candidate List Combination}\label{sec:calibration}
As the gold candidate recall decides the upper bound of an (X)EL system, candidate lists with close to 100\% recall are ideal. However, this is hard to achieve for most low-resource languages where existing candidate generation models only provide candidate lists with low recall (less than 60\%, as we show in Section \ref{comparision}). Further, combination of candidate lists retrieved by different models is non-trivial as the scores are not comparable among models. For example, scores of \textsc{WikiMention} have probabilistic interpretation while scores of \textsc{Pivoting} do not.

We propose a simple method to solve this problem: we convert scores without probabilistic interpretation to ones that are scaled to the zero-one simplex. Given mention $m_i$ and its top-$n$ candidate entity list $\Ev_{i}$ along with their scores $\mathbf{S}_i$, the re-calibrated scores are identified as:
\begin{equation}
    p_{i,j}=\frac{\exp(\gamma \times s_{i,j})}{\sum_{s_{i,k}\in \mathbf{S}_i}\exp(\gamma \times s_{i,k})}
\end{equation}
where $\gamma$ is a hyper-parameter that controls the peakiness of the distribution.  
After calibration, it is safe to combine prior scores with an average. 

\subsection{Feature Design}\label{feature}
Next, we introduce the feature set for our disambiguation model, including features inspired by previous work \cite{sil2016one, ganea2016probabilistic, pan2017cross}, as well as novel features specifically designed to tackle the low-resource scenario.
We intentionally avoid features that take source language context words into consideration, as these would be heavily reliant on $\mathbbm{W}_{\textrm{eng}}$ and $\mathbbm{M}$ and weaken the transferability of the model.
The formulation and resource requirements of unary and binary features are shown in the top and bottom halves of Table \ref{tab:1} respectively.



For unary features, we consider the number of mentions an entity is related to as $f_l^3$, where we consider the entity $e_{i,j}$ related to mention $m_k$ if it co-occurs with any candidate entity of $m_k$~\cite{moro2014entity}. We also add the entity prior score $f_l^2$ among the whole Wikipedia~\cite{yamada2017learning} to reflect the entity's overall salience. The exact match number $f_l^4$ indicates mention coreference. 

For binary features, we attempt to deal with the noise and sparsity inherent in the co-occurrence counts of $f_g^1$. To tackle noise, we calculate the smoothed Positive Pointwise Mutual Information (PPMI) \cite{church1990word,ganea2016probabilistic} between two entities as $f_g^2$, which robustly estimates how much more the two entities co-occur than we expect by chance.
To tackle sparsity, we incorporate English entity embeddings of \citet{yamada2017learning}, and calculate embedding similarity between two entities as $f_g^3$. Similar techniques have also been used by existing works~\cite{ganea2017deep, kolitsas-etal-2018-end}. We also add the hyperlink count $f_g^4$ between a pair of entities as, if entity $e_i$'s Wikipedia page mentions $e_j$, they are likely to be related.

We name our proposed feature set that includes all features listed in Table \ref{tab:1} as \textsc{Feat}.

\subsection{\textsc{Burn}: Feature Combination Model}\label{sec:burn}
With the growing number of features, we posit that a linear model with greedy entity pair selection (Section \ref{sec:base_inference}) is not expressive enough to take advantage of a rich feature set. \citet{yamada2017learning} use Gradient Boosted Regression Trees (GBRT; \citet{friedman2001greedy}) to combine features, but GBRTs do not allow for end-to-end training and thus constrain the flexibility of the model. \citet{ganea2016probabilistic,ganea2017deep} propose to use Loopy Belief Propagation (LBP; \citet{murphy1999loopy}) to estimate the global score (Equation \eqref{eq:base_g}) and use non-linear functions to combine local and global scores (Equation \eqref{burn:final}). However, BP is challenging to implement, and previous work has not attempted to combine more fine-grained features (e.g. unary feature $\mathbf{\Phi}(e_{i,j})$) non-linearly.  

Instead, we propose a \emph{belief update recurrent network} (\textsc{Burn}) that combines features in a non-linear and iterative fashion. Compared to existing work \cite{naradowsky2016represent, ganea2016probabilistic, ganea2017deep} as well as our base model, the advantages of \textsc{Burn} are: (1) it is easy to implement with existing neural network toolkits, (2) parameters can be learned end-to-end, (3) it considers non-linear combinations over more fine-grained features and thus has potential to fit more complex combination patterns, (4) it can model (distance) relations between mentions in the document. 


Given unary feature vector $\mathbf{\Phi}(e_{i,j})$ with $d_l$ features, \textsc{Burn} replaces the linear combination in Equation \eqref{eq:base_l} with two fully connected layers:
\begin{align*}
    s_l(e_{i,j}|D)&={\Wv_l^2}^T(\sigma({\Wv_l^1}^T\mathbf{\Phi}(e_{i,j}))) \\
    &+{\Wv_l^3}^T\mathbf{\Phi}(e_{i,j})
\end{align*}
where $\textbf{W}_l^1 \in \mathbbm{R}^{d_l \times h_l}$, $\textbf{W}_l^2 \in \mathbbm{R}^{h_l \times 1}$ and $\textbf{W}_l^3 \in \mathbbm{R}^{d_l \times 1}$. $\sigma$ is a non-linear function, for which we use leaky rectified linear units (Leaky ReLu; \citet{maas2013rectifier}). We add a linear addition of the input to alleviate the gradient vanishing problem. Equation \eqref{eq:base_e} is revised in a similar way. 

\begin{figure}[t]
    \centering
    \includegraphics[width=0.93\columnwidth]{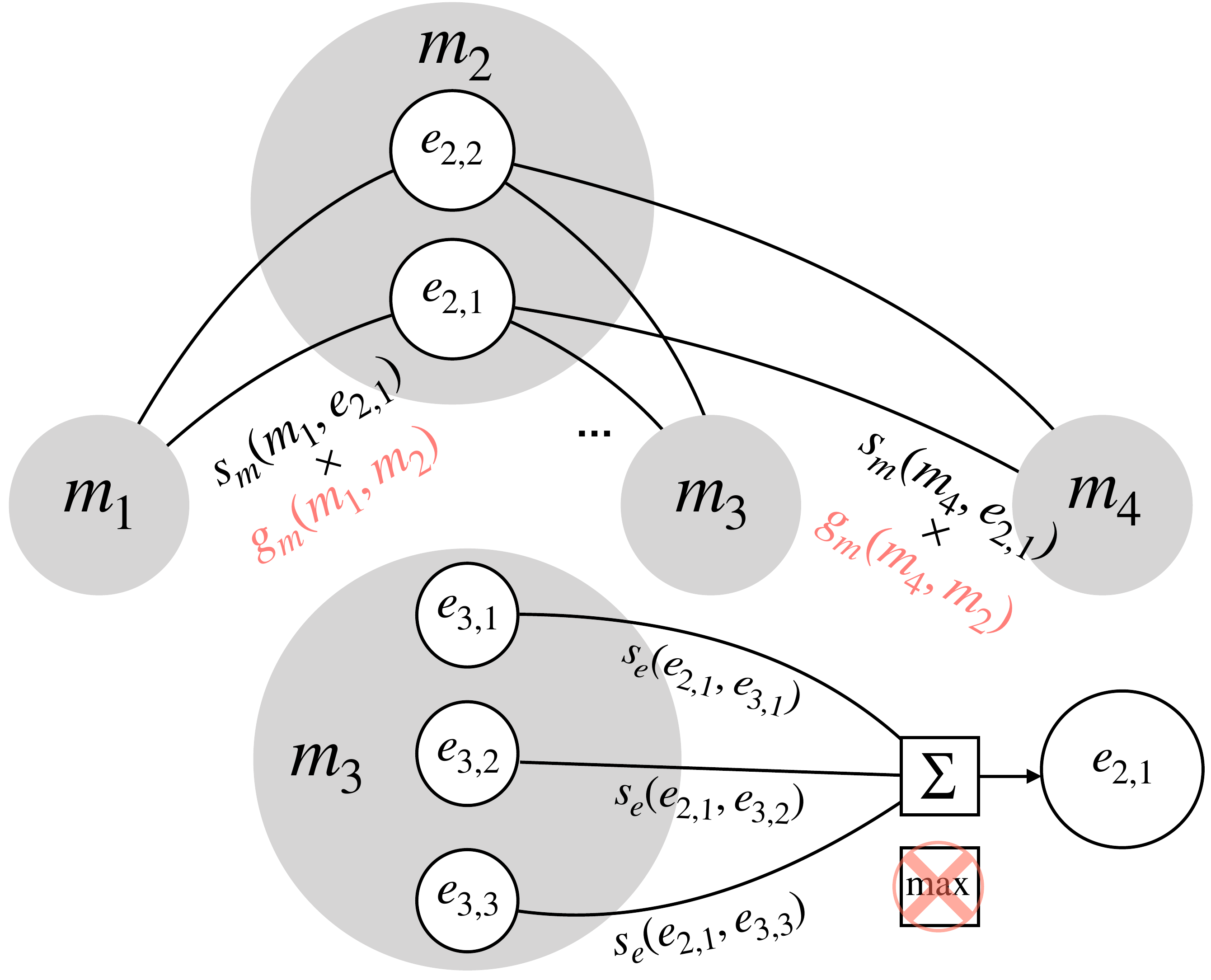}
    \caption{Top: the global score of an entity is a \emph{weighted} aggregation of mention evidence from context mentions, instead of an average. Bottom: each mention evidence is a \emph{weighted} entity-pair score, instead of the max.}
    \label{fig:change}
\end{figure}

As discussed in Equation \eqref{eq:base_m}, our baseline model calculates the mention evidence greedily. However, there may be many candidate entities for each mention, some containing noise. 
\textsc{Burn} solves this problem by weighting $s_e(e_{i,j}, e_{k,w})$ with the current entity probability $p(e_{k,w}|D)$. An illustration is in the bottom of Figure \ref{fig:change}.
The evidence from $m_k$ is now defined as:
\begin{equation}\label{eq:mention_semantic_related}
 s_m(m_k, e_{i, j}) = \sum_{w=1}^{|C_k|}s_e(e_{i, j}, e_{k, w})p(e_{k, w}|D)   
\end{equation}

Instead of simply averaging mention evidence in Equation \eqref{eq:base_g}, we also use a gating function to control the influence of $m_k$'s mention evidence on $m_i$ (top of Figure \ref{fig:change}), giving score
\begin{equation*}\label{eq:global_score}
    s_g(e_{i,j}|D)=\sum_{k \neq i}g_m(m_i, m_k)s_m(m_k, e_{i, j})
\end{equation*}
The gating function $g$ is essentially a lookup table that has one scalar for each distance (in words) between two mentions. 
We train this table along with all other parameters of the model.
The motivation for this gating function is that a mention is more likely to be coherent with a nearby mention than a distant one. 
We assume that this is true for almost all languages, and thus will be useful even without training in the language to be processed. 

As shown in Equation \eqref{eq:mention_semantic_related}, there is a circular dependency between entities. To solve this problem, we iteratively update the probability of entities until convergence or reaching a maximum number of iterations $T$. In iteration $t$, the calculation of $s_m$ will use entity probabilities from iteration $t-1$. The revised Equation \eqref{eq:mention_semantic_related} is as follows:
\begin{equation*}
 s_m^{t}(m_k, e_{i, j}) = \sum_{w=1}^{|C_k|}s_e(e_{i, j}, e_{k, w})p^{t-1}(e_{k, w}|D)
\end{equation*}
Unrolling this network through iterations, we can see that this is in fact a recurrent neural network.  

\noindent
\textbf{Training \textsc{Burn}}: 
The weights of \textsc{Burn} are learned end-to-end with the objective function:
\begin{equation*}
    L(\mathcal{D}, \mathcal{E})=-\sum_{D\in \mathcal D}\sum_{m_i \in D}\log(p^{T}({e}_i^*|D)).
\end{equation*}
As discussed above, the disambiguation model is fully language-agnostic and it does not require any annotated EL data or other resources in the source language. The model weights $W_l$, $W_g$ and the lookup table $g_m$ of gating function are trained on the TAC-KBP 2010 English training set~\cite{ji2010overview} \emph{only} and used as-is in another language. We use TAC-KBP 2012 English test set \cite{mayfield2012overview} as our development set.

\section{Experiment \rom{2}: Improving Low-resource XEL}
\label{experiment}
\begin{table*}[t]
  \small
  \centering
  \begin{tabular}{cp{0.5cm}p{0.5cm}p{0.5cm}cccccc}
    \toprule
     Block Index & 
     $\mathbbm{W}_{\textrm{eng}}$ & $\mathbbm{W}_{\textrm{src}}$ & $\mathbbm{M}$ & Candidates & Inference & ti & om & rw & si  \\
     \midrule
     \multirow{7}{*}{1}&
     \multirow{7}{*}{\checkmark} & 
     \multirow{7}{*}{} & \multirow{7}{*}{} & \multirow{6}{*}{\textsc{Pivoting}} 
     & Gold Candidate Recall & 36.2 & 20.9 & 59.6 & 32.1 \\
     \cmidrule{6-10}
     & & & & &$p(e|m)$ &  32.9 & 18.2 & 54.9 & 11.8  \\  
     & & & & &\textsc{Base} + \textsc{Greedy}  & \underline{33.7} & \underline{18.5} & \underline{55.9} & \underline{20.5} \\  
     & & & & &\textsc{Feat} + \textsc{Greedy} & 33.7 & 13.6 & 46.2 & 15.5 \\  
     & & & & &\textsc{Base} + \textsc{Burn} &  \textbf{34.9} & \textbf{19.4} & \textbf{56.2} & \textbf{21.1}  \\  
     & & & & &\textsc{Feat} + \textsc{Burn}& 34.5 & 17.8 & 50.9 & 10.6  \\   
     \midrule
    \multirow{7}{*}{2} &
    \multirow{7}{*}{\checkmark} & \multirow{7}{*}{\checkmark} & \multirow{7}{*}{\checkmark} & \multirow{7}{*}{\textsc{WikiMention}}
     & Gold Candidate Recall  & 21.9 & 45.3 & 45.6 & 66.6 \\
     \cmidrule{6-10}
     & & & & & $p(e|m)$  &  21.5 & \textbf{41.0} & 45.1 & 63.1  \\
     & & & & &\textsc{Base} + \textsc{Greedy}  &  \cellcolor{Gray}\underline{\textbf{21.8}} & \cellcolor{Gray}\underline{38.4} & \cellcolor{Gray}\underline{44.9} & \cellcolor{Gray}\underline{64.4}  \\  
     & & & & &\textsc{Feat} + \textsc{Greedy} & 21.6 & 38.7 & 44.6 & 64.4   \\  
     & & & & &\textsc{Base} + \textsc{Burn} & \textbf{21.8} & 39.9 & 44.3 & \textbf{64.7}  \\  
     & & & & &\textsc{Feat} + \textsc{Burn} & \textbf{21.8} & 39.9 & \textbf{45.6} & \textbf{64.7}\\
     \midrule
     
     \multirow{7}{*}{3}& 
     \multirow{7}{*}{\checkmark} & \multirow{7}{*}{\checkmark} & \multirow{7}{*}{\checkmark} & \multirow{4}{*}{\textsc{WikiMention}}
     & Gold Candidate Recall  &  38.3 & 62.0 & 69.4 & 75.2 \\
     \cmidrule{6-10}
     & & & & & $p(e|m)$  &  33.6 & 54.0 & 66.0 & 66.8  \\
     & & & & &\textsc{Base} + \textsc{Greedy}  &  \underline{34.4} & \underline{53.3} & \underline{67.3} & \underline{68.1}  \\  
     & & & & \multirow{2}{*}{+ \textsc{Pivoting}} &\textsc{Feat} + \textsc{Greedy} & 34.5 & 50.3 & 57.8 & 67.2  \\  
     & & & & &\textsc{Base} + \textsc{Burn} & \textbf{35.6} & \textbf{54.5} & 65.2 & \textbf{70.3}  \\  
     & & & & &\textsc{Feat} + \textsc{Burn} & 35.2 & 53.6 & \textbf{67.5} & 68.8 \\
     \bottomrule
  \end{tabular}
  
    \caption{Accuracy (\%) of different systems. \checkmark shows the resource requirements. The performances of the end-to-end baseline system \colorbox{Gray}{grayed}. The performances of baseline disambiguation for each candidate generation model are \underline{underlined} and numbers in \textbf{bold} show the best performance for each setting. $p(e|m)$ refers to the method that chooses the highest prior score provided by corresponding candidate generation method. 
    }
  \label{tab:3}
\vspace{-5mm}
\end{table*}

Section \ref{base_experiment} demonstrated a dramatic performance degradation for XEL in realistic low-resource settings. In this section, we evaluate the utility of our proposed methods that improve low-resource XEL.

\subsection{Training Details}\label{train_detail}
All models are implemented in PyTorch~\cite{paszke2017automatic}. The size of the pre-trained entity embeddings~\cite{yamada2017learning} is 300, trained with a window size of 15 and 15 negative samples. The hidden size $h$ of both $\mathbf{W_l^1}$ and $\mathbf{W_g^1}$ is set to 128, the dropout rate is set to 0.5. For the gating function, we set mention distances that are larger than 50 tokens to 50, then bin the distances with a bin size of 4. We only consider the 30 nearest context mentions for each mention. The maximum number of iterations for inference is set to 20. We use the Adam optimizer with the default learning rate (\texttt{1e-3}) to train the model. The $\gamma$ of calibrated candidate combination is set to 1. It takes around two hours to train a \textsc{Greedy} model and ten hours to train a \textsc{Burn} model with a Titan X GPU, regardless of the feature set.
\subsection{Results}
Table \ref{tab:3} compares models on the datasets we introduce in Section \ref{base_experiment}. Given that the critical issue was the degradation of candidate recall of the resource-heavy \textsc{WikiMention} method in low-resource settings (Section \ref{base_experiment}), we first examine the alternative resource-light \textsc{Pivoting} model.
The first rows of block 1 and 2 of the table show the gold candidate recall of each method.
While \textsc{Pivoting} greatly exceeds \textsc{WikiMention} on \textit{ti}, which only has 168 Wikipedia pages, its performance is much lower on \textit{si}, which has 15k pages. Overall, while these two models could outperform each other in their respective favorable settings (when a similar pivot language exists for the former, and when a large Wikipedia exists for the latter), it is challenging to decide which is more appropriate in the face of the realistic setting of existent, but scarce, resources.

Thus, in the third block of the table we show results for the hybrid candidate generation model which uses both \textsc{WikiMention} and \textsc{Pivoting}.
Compared to \textsc{WikiMention}, this method improves the gold candidate recall between 9 to 24\% over all four low-resource languages. The improvement ($>$ 15\%) is especially considerable for \textit{om} and \textit{rw}.
This reflects the fact that there are a significant number of unique candidate entities retrieved by these two candidate generation methods, and developing a proper way to combine them together results in higher-quality candidate lists.  
Notably, this method has also increased the headroom for a disambiguation model to contribute -- in contrast to the \textsc{WikiMention} setting where the difference between prior $p(e|m)$ and gold accuracy was minimal, now there is a 3-9\% accuracy gap between the two settings.


Next, we turn to methods that close this gap.
Focusing on this third block of the table, we can see that the proposed disambiguation model can take advantage of better candidate lists and yields significantly better results on all four languages.
Notably, we observe that \textsc{Burn} consistently yields the best performance over all languages, improving by 0.2 to 3.3\% over \textsc{Greedy}. This result demonstrates the advantage of iterative non-linear feature combination in low-resource settings.
In contrast, there is not a consistent improvement from the proposed feature set \textsc{Feat} compared to the baseline \textsc{Base}.
This is interesting as \textsc{Feat}+\textsc{Burn} outperformed \textsc{Base}+\textsc{Burn} by more than 10\% on the English development set on which it was validated.
We suspect this is because the feature value distribution of the English training data is different from that of low-resource languages, leading to sub-optimal transfer.
We leave training algorithms for bridging this gap as an interesting avenue of future work. 

In the context of the end-to-end system, the combination of our proposed methods brings 6-23\% improvement over the baseline system. 
For languages (\textit{ti}, \textit{om}, \textit{rw}) where resources are relative scarce, the improvement is especially considerable, ranging from 13 to 23\%, indicating that our work is a promising first step towards improving XEL in realistic low-resource scenarios.  


\section{Conclusion}


This paper has made two major contributions to the study of low-resource cross-lingual entity linking (XEL).
First, we perform an extensive empirical evaluation on the effect of different resource availability assumptions on XEL and demonstrate that (1) the accuracy of existing systems greatly degrades on true low-resource settings, and (2) standard candidate generation systems constrain the performance of end-to-end XEL.
This fact has been under-discussed in existing work and we argue that more attention should be paid to candidate generation for low-resource XEL.
Second, based on our empirical study, we propose three methodologies for candidate generation and disambiguation that make the best use of limited resources we will have in realistic settings.
Experimental results suggest that our proposed methodologies are effective under extremely limited-resource scenarios, giving improvements in 6-23\% end-to-end linking accuracy over the baseline system.  

An immediate future focus is further improving the performance of candidate generation models in realistic low-resource settings.
Further, we could consider more sophisticated strategies for cross-lingual training of entity disambiguation systems that fill the gap between English training data and real world low-resource data.


\section{Acknowledgements}
We would like to thank the anonymous reviewers for their useful feedback. This material is based upon work supported in part by the Defense Advanced Research Projects Agency Information Innovation Office (I2O) Low Resource
Languages for Emergent Incidents (LORELEI) program under Contract No. HR0011-15-C0114.
The views and conclusions contained in this document are those of the authors and should not be
interpreted as representing the official policies, either expressed or implied, of the U.S. Government. The U.S. Government is authorized to reproduce and distribute reprints for Government
purposes notwithstanding any copyright notation
here on.

\bibliography{emnlp-ijcnlp-2019.bib}
\bibliographystyle{acl_natbib}


\end{document}


\begin{CJK*}{UTF8}{gbsn}
\maketitle

\appendix

\section{Training Details}

All models are implementented in the PyTorch Framework. The size of the pre-trained entity embeddings\footnote{https://wikipedia2vec.github.io/wikipedia2vec/\\pretrained/} is 300, trained with a window size set to 15 and negative sample set to 15.  The hidden state of both unary features and binary features are set to 128, the dropout rate is set to 0.5. 

For the gating function, we set mention distances that are larger than 50 tokens are to 50, then bin the distance with a bin size of 4. We use a lookup table to find the corresponding gating parameter. We only consider the 30 nearest context mentions for one mention. The maximum number of iterations for inference is set to 20.

We use Adam optimizer with the default learning rate (\texttt{1e-3}) to train the model. The model is trained on the TAC-KBP 2010 English training set~\cite{ji2010overview} and hyper-parameters are tuned on the TAC-KBP 2012 English evaluation dataset~\cite{mayfield2012overview}.
In the candidate generation process, we set $K=30$ for all languages.

The $\gamma$ in Section 4.1 is set to 1.
\section{\textsc{Pivoting} Candidate Generation}

We select training languages for the pivoting model based on language family similarity with the source language in question. The models are tuned on a validation set from Wikipedia, which has no overlap with our DARPA test datasets.
The training languages we choose are -- Amharic (for Tigrinya), Somali (for Oromo), Kirundi (for Kinyarwanda) and Hindi (for Sinhala)~\cite{rijhwani19aaai}. 

We train the models on bilingual entity maps obtained from Wikipedia language links, on the transfer language \emph{only}, to maximize cosine similarity between entity encodings. No data in the source language is used for training.




\bibliography{emnlp-ijcnlp-2019}
\bibliographystyle{acl_natbib}

\end{CJK*}